\documentclass[12pt,a4paper]{article}
\usepackage[dvips]{color,graphicx}
\usepackage{amsmath,amssymb}
\usepackage{epsf}
\usepackage{mlapa}

\newtheorem{theorem}{Theorem}

\newtheorem{corollary}[theorem]{Corollary}

\newtheorem{definition}[theorem]{Definition}

\title{Asymptotic Accuracy of Distribution-Based Estimation
for Latent Variables}

\author{Keisuke Yamazaki\\
       k-yam@math.dis.titech.ac.jp \\
       Department of Computational Intelligence and Systems Science,\\
       Tokyo Institute of Technology\\
       G5-19 4259 Nagatsuta Midori-ku Yokohama, Japan
	}
\date{}
\begin{document}
\sloppy
\maketitle

\begin{abstract}
Hierarchical statistical models are widely employed in information science and data engineering.
The models consist of two types of variables: observable variables that represent the given data and latent variables for the unobservable labels.
An asymptotic analysis of the models plays an important role in evaluating the learning process;
the result of the analysis is applied not only to theoretical but also to practical situations, such as
optimal model selection and active learning.
There are many studies of generalization errors, which
measure the prediction accuracy of the observable variables.
However, the accuracy of estimating the latent variables has not yet been elucidated.
For a quantitative evaluation of this,
the present paper formulates distribution-based functions for the errors in the estimation of the latent variables.
The asymptotic behavior is analyzed for both the maximum likelihood and the Bayes methods.
\end{abstract}

\section{Introduction}
\sloppy
Hierarchical probabilistic models, such as mixture models, are mainly
employed in unsupervised learning.
The models have two types of variables: observable and latent.
The observable variables represent the given data, and the latent ones
describe the hidden data-generation process.
For example, in mixture models that are employed for clustering tasks, observable variables are the attributes of the given data and the latent ones are the
unobservable labels.

One of the main concerns in unsupervised learning is the analysis of the hidden processes,
such as how to assign clustering labels based on the observations.
Hierarchical models have an appropriate structure for this analysis,
because it is straightforward to estimate the latent variables from the observable ones.
Even within the limits of the clustering problem,
there are a great variety of ways to detect unobservable labels, both probabilistically and deterministically,
and many criteria have been proposed to evaluate the results  \cite{Dubes+Jain}.
For parametric models, the focus of the present paper,
learning algorithms such as the expectation-maximization (EM) algorithm
and the variational Bayes (VB) method \cite{Attias99,Ghahramani00,Smidl05,Beal03}
have been developed for estimating the latent variables.
These algorithms must estimate both the parameter and the variables,
since the parameter is also unknown in the general case.

Theoretical analysis of the models plays an important role in evaluating
the learning results.
There are many studies on predicting performance in situations where both training and test data are described by the observable variables.
The results of asymptotic analysis have been used for practical applications, such as model selection
and active learning \cite{Akaike,Takeuchi1976,book:Fedorov:1972}.
The simplest case of the analysis is when the learning model contains the true model,
which generates the data.
Recently,
it has been pointed out that
when there is the redundant range/dimension of the latent variables in the learning model,
singularities exist in the parameter space and the conventional statistical analysis is not valid \cite{Amari}.
To tackle this issue,
a theoretical analysis of the Bayes method was established using
algebraic geometry \cite{Watanabe09:book}.
The generalization performance was then derived
for various models \cite{Yamazaki03a,Yamazaki03b,Rusakov,Aoyagi10,Zwiernik11}.
Based on this analysis of the singularities, some criteria for model selection
have been proposed \cite{watanabe_WAIC,yamazaki05_SingIC,yamazaki06_SingIC}.

Although validity of the learning algorithms is necessary for unsupervised tasks, 
statistical properties of the accuracy of the estimation of the latent variables have not been studied sufficiently.
\begin{table}[t]
\centering
\caption{Estimation classification according to the target variable and the model case}
\begin{tabular}{|c|c|c|}
\hline
Estimation Target \textbackslash Model Case & Regular Case & Singular Case \\
\hline
Observable Variable & Reg-OV estimation  & Sing-OV estimation \\
\hline
Latent Variable & Reg-LV estimation & Sing-LV estimation \\
\hline
\end{tabular}
\label{tab:4est}
\end{table}
Table \ref{tab:4est} summarizes the classification according to the target variable of estimation
and the model case.
We will use the abbreviations shown in the table to specify the target variable and the model case;
for example, Reg-OV estimation stands for estimation of the observable variable in the regular case.
As mentioned above, theoretical analysis have been conducted in both the Reg-OV and the Sing-OV estimations.
On the other hand, there is no statistical approach to measure the accuracy of 
the Reg-LV or the Sing-LV estimation.

The goal of the present paper is to provide an error function for measuring the accuracy,
which is suitable for the unsupervised learning with hierarchical models,
and to derive its asymptotic form.
For the first step, we consider the simplest case, in which the
attributes, such as the range and dimension, of the latent variables are known;
there is no singularity in the parameter space.
This corresponds to the Reg-OV estimation in the table.
Since the mathematical structure of the parameter is much more complicated in the singular case,
we leave the analysis of the Sing-LV estimation for \cite{Yamazaki12}.
The main contributions of the present paper are the following three items:
(1) estimation for the latent variables falls into three types as shown in Fig.~\ref{fig:PredEst}
and their error functions are formulated in a distribution-based manner;
(2) the asymptotic forms of the error functions are derived on the maximum likelihood
and the Bayes methods in Type I and variants of Types II and III shown in Fig.~\ref{fig:vTypes};
(3) it is determined that the Bayes method is more accurate than the maximum likelihood method
in the asymptotic situation.

The rest of this paper is organized as follows:
In Section \ref{sec:Prediction} we explain the estimation of latent variables by comparing it with the prediction of observable variables.
In Section \ref{sec:definitions} we provide the formal definitions of the estimation methods and the error functions.
Section \ref{sec:analysis} then presents the main results for the asymptotic forms and the proofs.
Discussions and conclusions are stated in Sections \ref{sec:disc} and \ref{sec:conc}, respectively.
\section{Estimations of Variables}
\label{sec:Prediction}
This section distinguishes between the estimation of latent variables and the prediction of observable variables.
There are variations on the estimation of latent variables due to the estimated targets.

Assume that the observable data and unobservable labels are represented
by the observable variables $x$ and  the latent variables $y$, respectively.
Let us define that $x\in R^M$ and $y\in \{1,2,\dots,K\}$.
In the case of a discrete $x$ such as $x\in\{1,2,\dots,M\}$,
all the results in this paper hold if $\int dx$ is replaced with $\sum_{x=1}^M$.
A set of $n$ independent data pairs is expressed as $(X^n,Y^n)=\{(x_1,y_1),\dots,(x_n,y_n)\}$,
where $X^n=\{x_1,\dots,x_n\}$ and $Y^n=\{y_1,\dots,y_n\}$.
More precisely, there is no dependency between $x_i$ and $x_j$
or between $y_i$ and $y_j$ for $i\neq j$.
\begin{figure}[t]
\begin{tabular}{c|c}
\begin{minipage}{0.5\hsize}
\centering
\includegraphics[angle=-90,width=\columnwidth]{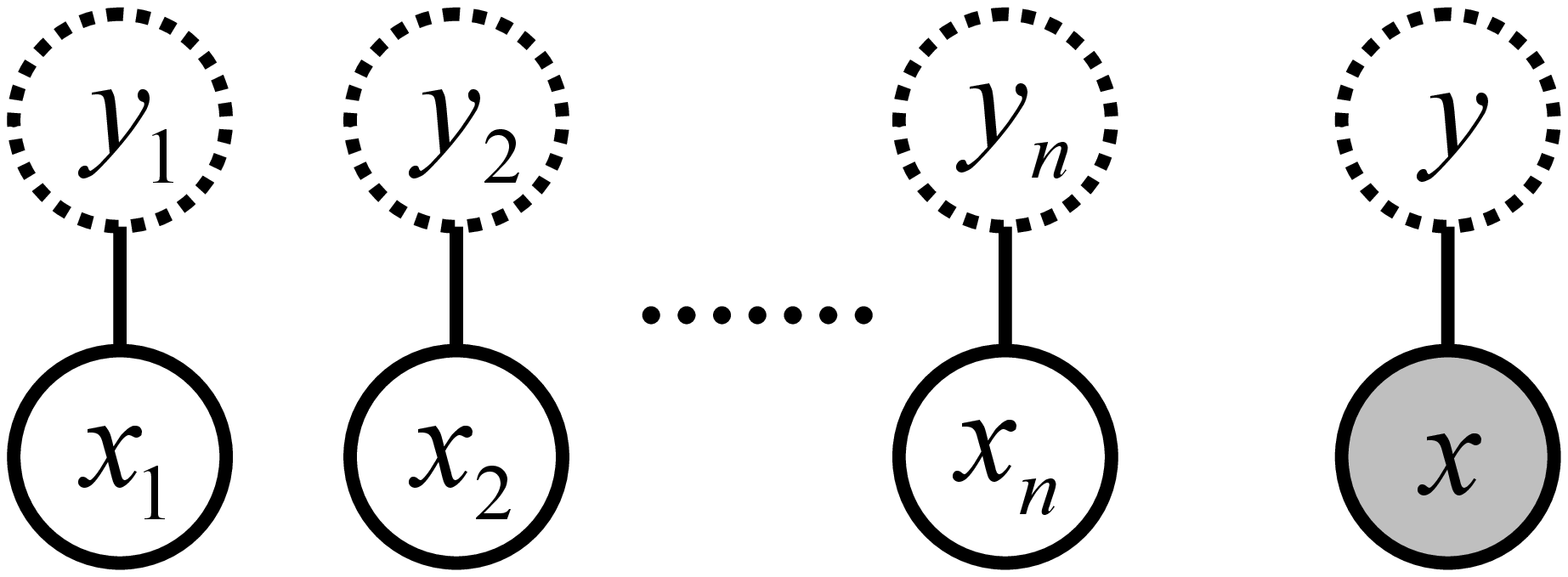}
\text{Prediction of observable variables}
\label{fig:regfig1}
\end{minipage}
&
\begin{minipage}{0.5\hsize}
\centering
\includegraphics[angle=-90,width=\columnwidth]{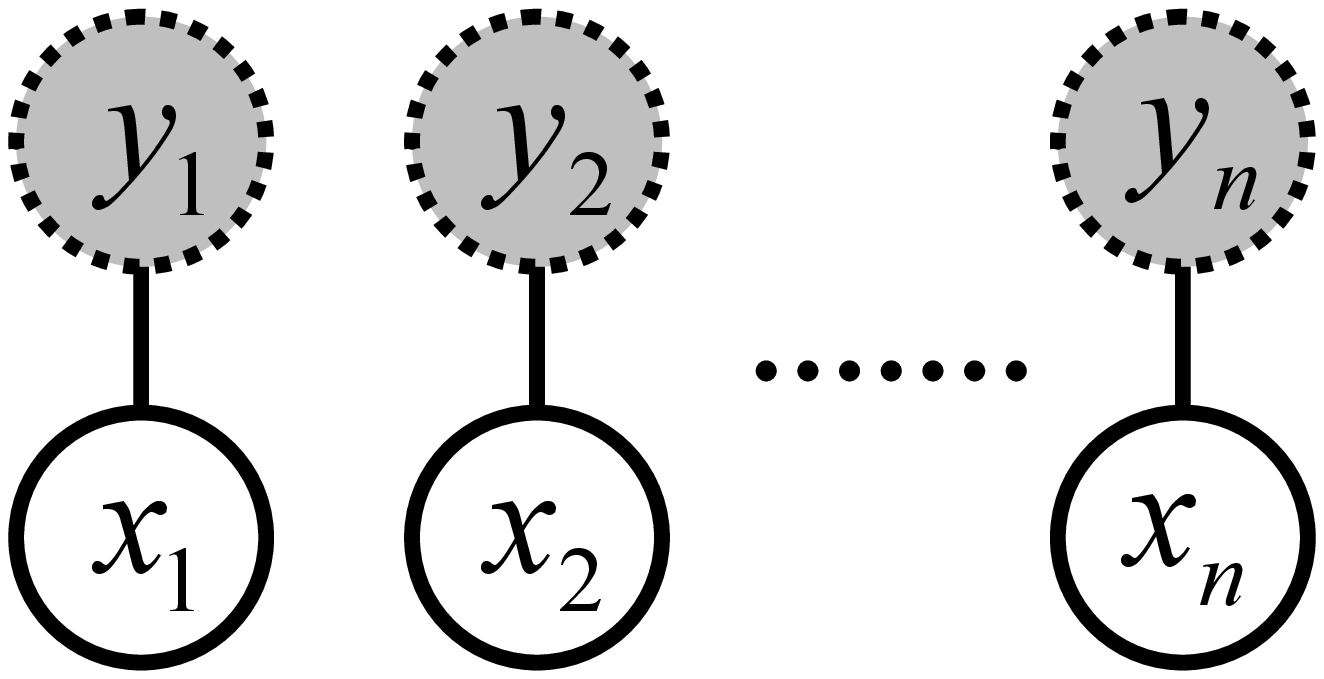}
\text{Type I}
\label{fig:regfig2}
\end{minipage}
\\
\hline\\
\begin{minipage}{0.5\hsize}
\centering
\includegraphics[angle=-90,width=\columnwidth]{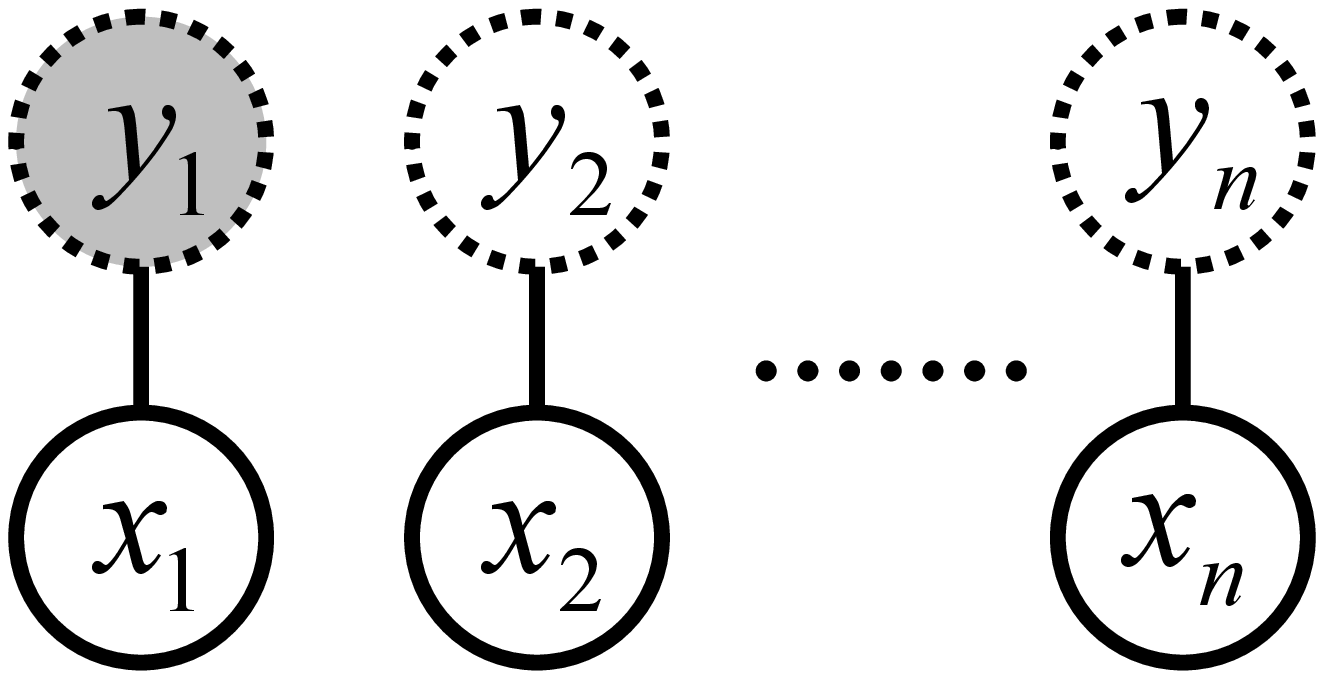}
\text{Type II}
\label{fig:regfig3}
\end{minipage}
&
\begin{minipage}{0.5\hsize}
\centering
\includegraphics[angle=-90,width=\columnwidth]{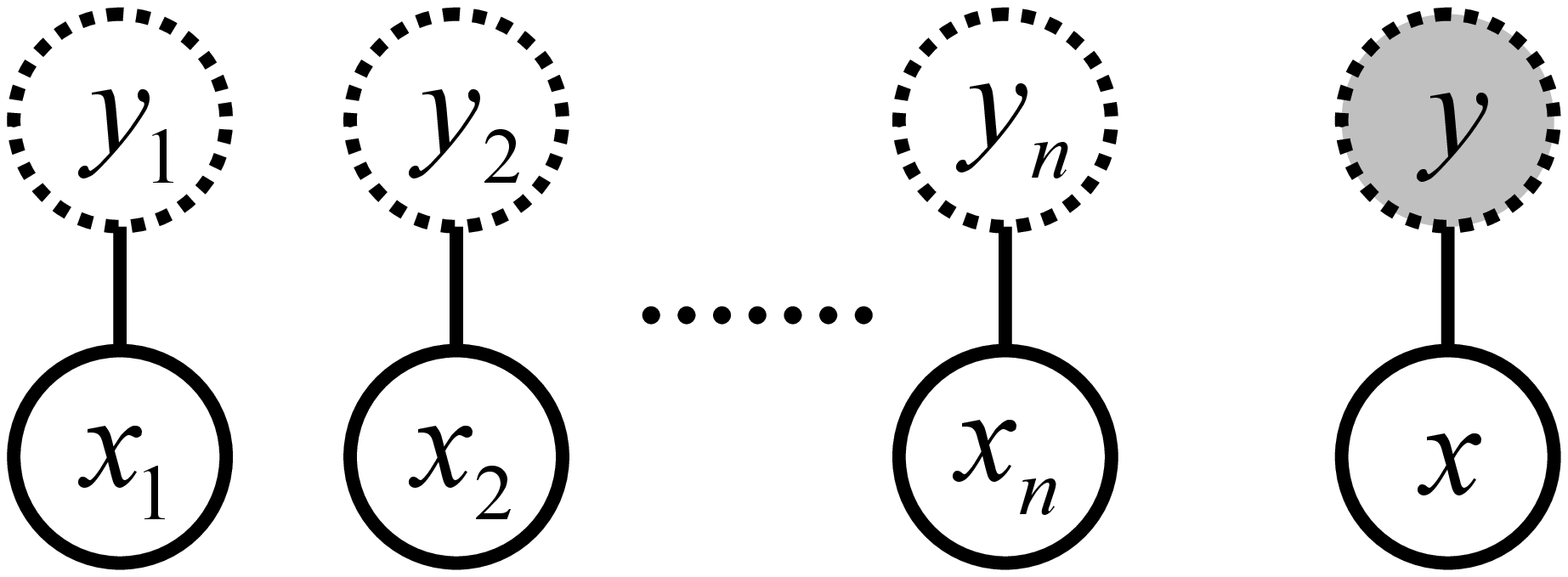}
\text{Type III}
\label{fig:regfig4}
\end{minipage}
\end{tabular}
\caption{Prediction of observable variables and estimations of latent variables.
The observable data are $\{x_1,\dots,x_n\}$. Solid and dotted nodes are observable and unobservable, respectively.
Gray nodes are estimation targets.}
\label{fig:PredEst}
\end{figure}

Figure \ref{fig:PredEst} shows a variety of estimations of variables:
prediction of an observable variable and three types of estimations of latent variables.
Solid and dotted nodes are the observable and latent variables, respectively.
A data pair is depicted by a connection between two nodes.
The gray nodes are the target items of the estimations.
We consider a stochastic approach, where the probability distribution of the target(s)
is estimated from the training data $X^n$.

The top-left panel shows the prediction of unseen observable data.
Based on $X^n$, the next observation $x=x_{n+1}$ is predicted.
The top-right panel shows the estimation of $Y^n$,
which is referred to as Type I.
In the stochastic approach, the joint probability of $Y^n$ is estimated.
The bottom-left panel shows marginal estimation, referred to as Type II.
The marginal probability of $y_i$ ($y_1$ is the example in the figure) is estimated;
the rest of the latent variables in the probability are marginalized out.
Note that there is no unseen/future data in either of Types I or II.
The bottom-right panel shows estimation of $y$ in the unseen data, which is referred to as Type III.
The difference between this and Type II is the training data;
the corresponding observable part of the target is included in the training set in Type II, but it is not included in Type III.
In the present paper we use a distribution-based approach to analyze the theoretical accuracy of a Type-I estimation, but we also consider connections to the other types.
\section{Formal Definitions of Estimation Methods and Accuracy Evaluations}
\label{sec:definitions}
This section presents the maximum likelihood and Bayes methods for estimating latent variables and the corresponding error functions.
Here, we consider only the Type-I estimation problem for the joint probability of the hidden part.
The other types will be defined and discussed in Section \ref{sec:disc}.

Let $p(x,y|w)=p(y|w)p(x|y,w)$ be a learning model, where $w\in W\subset R^d$ is the parameter.
The probability of the observable data is expressed as
\begin{align*}
p(x|w)=&\sum_{y=1}^K p(y|w)p(x|y,w).
\end{align*}
Assume that the true model generating the data $(X^n,Y^n)$ is expressed as
$q(x,y)=p(y|w^*)p(x|y,w^*)$, where $w^*$ is the true parameter,
and that the following Fisher information matrices exist and are positive definite;
\begin{align*}
\{ I_{XY}(w^*) \}_{ij} =& E\biggl[ \frac{\partial \ln p(x,y|w^*)}{\partial w_i}
\frac{\partial \ln p(x,y|w^*)}{\partial w_j}\biggr],\\
\{ I_{X}(w^*) \}_{ij} =& E\biggl[ \frac{\partial \ln p(x|w^*)}{\partial w_i}
\frac{\partial \ln p(x|w^*)}{\partial w_j}\biggr],\\
\end{align*}
where the expectation is
\begin{align*}
E[f(x,y)]=\int \sum_{y=1}^{K} f(x,y)p(x,y|w^*)dx.
\end{align*}
This condition requires the identifiability of the true model,
i.e., $q(y)>0$ for all $y$ and $i\neq j\Rightarrow q(x|y=i)\neq q(x|y=j)$.
The joint probability distribution of $(X^n,Y^n)$ is denoted by $q(X^n,Y^n)=\prod_{i=1}^n q(x_i,y_i)$.

We introduce two ways to construct a probability distribution of $Y^n$ based on the observable $X^n$.
First, we define an estimation method based on the maximum likelihood estimator.
The likelihood is defined by
\begin{align*}
L_X(w) =& \prod_{i=1}^n p(x_i|w).
\end{align*}
The maximum likelihood estimator $\hat{w}_X$ is given by
\begin{align*}
\hat{w}_X =& \arg\max L_X(w).
\end{align*}
\begin{definition}[The maximum likelihood method]
In the maximum likelihood estimation, 
the estimated distribution of the latent variables
is defined by
\begin{align}
p(Y^n|X^n) =& \frac{p(X^n,Y^n|\hat{w}_X)}{\sum_{Y^n}p(X^n,Y^n|\hat{w}_X)} \nonumber\\
=& \prod_{i=1}^n \frac{p(x_i,y_i|\hat{w}_X)}{\sum_{y_i}p(x_i,y_i|\hat{w}_X)} = \prod_{i=1}^n p(y_i|x_i,\hat{w}_X).\label{eq:MLmethod}
\end{align}
\end{definition}
The notation $p(Y^n|X^n,\hat{w}_X)$ is used when the method is emphasized.

Next, we define the Bayesian estimation.
Let the likelihood of the joint probability distribution be
\begin{align*}
L_{XY}(w) =& \prod_{i=1}^n p(x_i,y_i|w).
\end{align*}
The marginal likelihood functions are given by
\begin{align*}
Z(X^n,Y^n) =& \int L_{XY}(w)\varphi(w;\eta)dw,\\
Z(X^n) =& \sum_{Y^n}Z(X^n,Y^n)=\int L_X(w) \varphi(w;\eta)dw,
\end{align*}
where $\varphi(w;\eta)$ is a prior with the hyperparameter $\eta$.
We assume that the support of the prior includes $w^*$.
\begin{definition}[The Bayes method]
In the Bayes estimation, the estimated distribution of $Y^n$ is expressed as
\begin{align}
p(Y^n|X^n) =& \frac{Z(X^n,Y^n)}{Z(X^n)}.\label{eq:Bayesmethod}
\end{align}
Based on the posterior distribution defined by
\begin{align*}
p(w|X^n) = \frac{1}{Z(X^n)}L_X(w)\varphi(w;\eta),
\end{align*}
the estimated distribution has another equivalent form
\begin{align}
p(Y^n|X^n) =& \int \prod_{i=1}^n p(y_i|x_i,w)p(w|X^n)dw.\label{eq:Bayesmethod2}
\end{align}
\end{definition}
Comparing Eq.~\ref{eq:Bayesmethod2} with Eq.~\ref{eq:MLmethod}
reveals that the Bayes estimation is based on the expectation over the posterior
instead of the plug-in parameter $\hat{w}_X$.

The distribution of $Y^n$ in the true model is uniquely expressed as
\begin{align*}
q(Y^n|X^n)=\prod_{i=1}^n q(y_i|x_i) =\prod_{i=1}^n \frac{q(x_i,y_i)}{q(x_i)},
\end{align*}
where $q(x_i)=\sum_{y_i=1}^{K} q(x_i,y_i)$.
Accuracy of the latent variable estimation is measured by the
difference between  the true distribution $q(Y^n|X^n)$ and the estimated one $p(Y^n|X^n)$.
For the present paper, we define the error function as the average Kullback-Leibler divergence,
\begin{align}
D(n) =& \frac{1}{n}E_{X^n}\biggl[ \sum_{Y^n}q(Y^n|X^n)\ln\frac{q(Y^n|X^n)}{p(Y^n|X^n)}\biggr],\label{eq:deferror}
\end{align}
where the expectation is
\begin{align*}
E_{X^n}[f(X^n)] =& \int f(X^n) q(X^n)dX^n.
\end{align*}
Note that this function is available for any construction of $p(Y^n|X^n)$
when we consider the cases of the maximum likelihood and the Bayes methods
below.
\section{Asymptotic Analysis of the Error Function}
\label{sec:analysis}
In this section we present the main theorems for the asymptotic forms of the error function.
\subsection{Asymptotic Errors of the Two Methods}
In the unsupervised learning, there is label switching,
which makes interpretation of the estimation result difficult.
For example, define the parameter $w^*_s$ as $p(x,y=1|w^*_s)=p(x,y=2|w^*)$,
$p(x,y=2|w^*_s)=p(x,y=1|w^*)$, and $p(x,y=k|w^*_s)=p(x,y=k|w^*)$ for $k>2$.
In this parameter, the label $y=1$ and $y=2$ are switched compared with $w^*$.
It holds that $p(x|w^*_s)=p(x|w^*)$ whereas $p(x,y|w^*_s)\ne p(x,y|w^*)$.
Therefore, the estimation methods can search for $w^*_s$ as the true parameter instead of $w^*$
since there is no information of the true labels.
In the present paper, we focus on the best performance, where we successfully estimate the true parameter.
In other words, we define the true parameter according to the estimated label assignment.
Under the best performance situation,
the maximum likelihood estimator $\hat{w}_X$ converges to $w^*$ in probability,
and the posterior distribution of the Bayes method converges to the normal distribution,
the mean of which is $\hat{w}_X$, in law.
Then, it is obvious that the error function $D(n)$ goes to zero at $n\rightarrow \infty$.

The following theorems show the speed of decrease of the error function;
\begin{theorem}[The asymptotic error of the maximum likelihood method]
\label{th:asympMLerror}
In the latent variable estimation given by Eq.\ref{eq:MLmethod},
the error function Eq.\ref{eq:deferror} has the following asymptotic form:
\begin{align*}
D(n) =& \frac{1}{2n}\mathrm{Tr}[\{I_{XY}(w^*)-I_X(w^*)\}I^{-1}_X(w^*)] + o\bigg(\frac{1}{n}\bigg).
\end{align*}
\end{theorem}
\begin{theorem}[The asymptotic error of the Bayes method]
\label{th:asympBayeserror}
In the latent variable estimation given by Eq.\ref{eq:Bayesmethod},
the error function Eq.\ref{eq:deferror} has the following asymptotic form:
\begin{align*}
D(n) =& \frac{1}{2n}\ln \det \big[I_{XY}(w^*)I^{-1}_X(w^*)\big] + o\bigg(\frac{1}{n}\bigg).
\end{align*}
\end{theorem}
The proofs are in the appendix.
The dominant order is $1/n$ in both methods,
and its coefficient depends on the Fisher information matrices.
It is not an unaccountable result that the error value depends on the position of $w^*$.
For example, let us consider cluster analysis
and assume that distances among the clusters are large.
Since we can easily distinguish the clusters,
there is not much additional information on the label $y$.
Then, $I_{XY}(w^*)$ is close to $I_X(w^*)$, which makes $D(n)$ small in both methods.
The true parameter generally determines difficulty of tasks in the unsupervised learning,
and the theorems reflect this fact.
We will present a more detailed discussion on the coefficient in Section \ref{sec:disc}.

The following corollary shows the advantage of the Bayes estimation.
\begin{corollary}
\label{cor:comp}
Let the error functions for the maximum likelihood and the Bayes methods be denoted by
$D^{\text{ML}}(n)$ and $D^{\text{Bayes}}(n)$, respectively.
Assume that $I_{XY}(w^*)\neq I_X(w^*)$.
For any true parameter $w^*$, there exists a positive constant $c$ such that
\begin{align*}
D^{\text{ML}}(n) - D^{\text{Bayes}}(n) \ge \frac{c}{n} + o\bigg(\frac{1}{n}\bigg).
\end{align*}
\end{corollary}
The proof is in the appendix.
This result shows that $D^{\text{ML}}(n) > D^{\text{Bayes}}(n)$
for a sufficiently large data size $n$.
\section{Discussion}
\label{sec:disc}
\subsection{Relation to Other Error Functions}
We now formulate the predictions of observable data and the remaining estimations for Types II and III,
and we consider the relations of their error functions to that of Type I.

First, we compare the Reg-LV estimation with the Reg-OV estimation.
In the observable-variable estimation,
the error function is referred to as the generalization error,
which measures the prediction performance on unseen observable data.
The generalization error is defined as
\begin{align*}
D_x(n) =& E_{X^n}\bigg[ \int q(x) \ln\frac{q(x)}{p(x|X^n)}dx\bigg],
\end{align*}
where $x$ is independent of $X^n$ in the data-generating process of $q(x)$.
The predictive distribution $p(x|X^n)$ is constructed by
\begin{align*}
p(x|X^n) =& p(x|\hat{w}_X)
\end{align*}
for the maximum likelihood method and
\begin{align*}
p(x|X^n) =& \int p(x|w)p(w|X^n)dw
\end{align*}
for the Bayes method.
Both methods estimation have the same dominant terms in their asymptotic forms,
\begin{align*}
D_x(n) =& \frac{d}{2n} + o\bigg(\frac{1}{n}\bigg).
\end{align*}
The coefficient of the asymptotic generalization error
depends only on the dimension of the parameter for any model,
but that of $D(n)$ is determined by both the model expression and the true parameter $w^*$.
This dependency appears when the learning model does not contain the true model in the Reg-OV estimation,
and $\hat{w}_X$ is used for approximation of the error function for model selection \cite{Takeuchi1976}
and active learning \cite{book:Fedorov:1972}.
In the same way, by replacing $w^*$ with $\hat{w}_X$,
Theorems \ref{th:asympMLerror} and \ref{th:asympBayeserror} enable us to calculate the error function
in the Reg-LV estimation.

In the observable-variable estimation,
the error $D_x(n)$ is approximated by the cross-validation and bootstrap methods
since unseen data $x_{n+1}$ are interchangeable with one of the given observable data.
On the other hand, there is no substitution for the latent variable,
which means that any numerical approximation does not exist for $D(n)$ in principle.
The theoretical results in the present paper are thus far the only way to estimate the accuracy.

Next, we discuss Type-II estimation; we focus on the value $y_i$ from $Y^n$ and its estimation accuracy.
Based on the joint probability, the estimation of $y_i$ is defined by
\begin{align*}
p(y_i|X^n) =& \sum_{Y^n \backslash y_i}p(Y^n|X^n),
\end{align*}
where the summation is taken over $Y^n$ except for $y_i$.
Thus the error function depends on which $y_i$ we exclude.
In order to measure the average effect of the exclusions, we define the error as follows:
\begin{align*}
D_{y|X^n}(n) =& E_{X^n}\bigg[\frac{1}{n}\sum_{i=1}^n \sum_{y_i}q(y_i|x_i)\ln \frac{q(y_i|x_i)}{p(y_i|X^n)}\bigg].
\end{align*}
The maximum likelihood method has the following estimation,
\begin{align*}
p(y_i|X^n) =& \sum_{Y^n\backslash y_i}\prod_{i=1}^n \frac{p(x_i,y_i|\hat{w}_X)}{p(x_i|\hat{w}_X)}\nonumber\\
=& \frac{p(x_1|\hat{w}_X)\cdots p(x_{i-1}|\hat{w}_X) p(x_i,y_i|\hat{w}_X) p(x_{i+1}|\hat{w}_X)\cdots p(x_n|\hat{w}_X)}{\prod_{i=1}^n p(x_i|\hat{w}_X)}\nonumber\\
=& \frac{p(x_i,y_i|\hat{w}_X)}{p(x_i|\hat{w}_X)}=p(y_i|x_i,\hat{w}_X).
\end{align*}
We can easily find that
\begin{align*}
D_{y|X^n}(n) =& E_{X^n}\bigg[\frac{1}{n}\sum_{i=1}^n \sum_{y_i=1}^{K}
q(y_i|x_i)\ln \frac{q(y_i|x_i)}{p(y_i|x_i,\hat{w}_X)}\bigg]\\
=& \frac{1}{n}E_{X^n}\bigg[ \sum_{Y^n}q(Y^n|X^n)\ln\frac{q(Y^n|X^n)}{p(Y^n|X^n,\hat{w}_X)}\bigg].
\end{align*}
Therefore, it holds that $D_{y|X^n}(n)=D(n)$ in the maximum likelihood method.
However, the Bayes method has the estimation,
\begin{align*}
p(y_i|X^n) =& \frac{\int p(x_1|w)\cdots p(x_{i-1}|w)p(x_i,y_i|w)p(x_{i+1}|w)\cdots p(x_n|w)\varphi(w;\eta)dw}{Z(X^n)},
\end{align*}
which indicates $D_{y|X^n}(n)\neq D(n)$.
A sufficient condition for $D_{y|X^n}(n)=D(n)$ is to satisfy $p(Y^n|X^n) = \prod_{i=1}^n p(y_i|X^n)$.

Finally, we consider the Type-III estimation.
The error is defined by
\begin{align*}
D_{y|x}(n) =& E_{X^n}\bigg[ \int q(x) \sum_{y=1}^{K}q(y|x)\ln\frac{q(y|x)}{p(y|x,X^n)}dx\bigg].
\end{align*}
Note that the new observation $x$ is not used for estimation of $y$,
or $D_{y|x}(n)$ will be equivalent to the Type-II error $D_{y|X^{n+1}}(n+1)$.
The maximum likelihood estimation $p(y|x,X^n)$ is given by
\begin{align*}
p(y|x,X^n) =& \frac{p(x,y|\hat{w}_X)}{p(x|\hat{w}_X)},
\end{align*}
and for the Bayes method it is
\begin{align}
p(y|x,X^n) =& \int \frac{p(x,y|w)}{p(x|w)}p(w|X^n)dw. \label{eq:BayesTypeIII}
\end{align}
Using the result in \cite{ShimodairaPDOI} for a variant Akaike information criterion (AIC) from partially observed data,
we immediately obtain the asymptotic form of $D_{y|x}(n)$ as
\begin{align*}
D_{y|x}(n) =& \frac{1}{2n}\mathrm{Tr}\bigg[ \bigg\{I_{XY}(w^*)-I_X(w^*)\bigg\}I_X(w^*)^{-1}\bigg] +o\bigg(\frac{1}{n}\bigg).
\end{align*}
We thus conclude that all estimation types have the same accuracy in the maximum likelihood method.
The difference of the training data between Types II and III does not asymptotically affect
the estimation results.
The analysis of the Type-III estimate in the Bayes method is left for future study.
\subsection{Variants of Types II and III}
\begin{table}[t]
\centering
\caption{Coefficients of the dominant order $1/n$ in the error functions}
\begin{tabular}{|c|c|c|c|c|}
\hline
& Prediction & Type I & Type II & Type III \\
\hline
ML & $d/2$ & $\mathrm{Tr}[\{I_{XY}-I_X\}I_X^{-1}]/2$ & $\mathrm{Tr}[\{I_{XY}-I_X\}I_X^{-1}]/2$ & $\mathrm{Tr}[\{I_{XY}-I_X\}I_X^{-1}]/2$ \\
\hline
Bayes & $d/2$ & $\ln\det[I_{XY}I_X^{-1}]/2$ & unknown & unknown\\
\hline
\end{tabular}
\label{tab:errors}
\end{table}
Table \ref{tab:errors} summarizes the results in the previous subsection.
The rows indicate the maximum likelihood (ML) and the Bayes methods, respectively.
The Fisher information matrices $I_{XY}(w^*)$ and $I_X(w^*)$ are abbreviated
in a form that does not include the true parameter, i.e., $I_{XY}$ and $I_X$.
The error functions of Types II and III in the Bayes method are still unknown.
The analysis is not straightforward when there is a single target of estimation, because the asymptotic expansion is not available
when the number of target nodes is constant with respect to the training data size $n$.

\begin{figure}[t]
\begin{tabular}{c|c}
\begin{minipage}{0.5\hsize}
\centering
\includegraphics[angle=-90,width=\columnwidth]{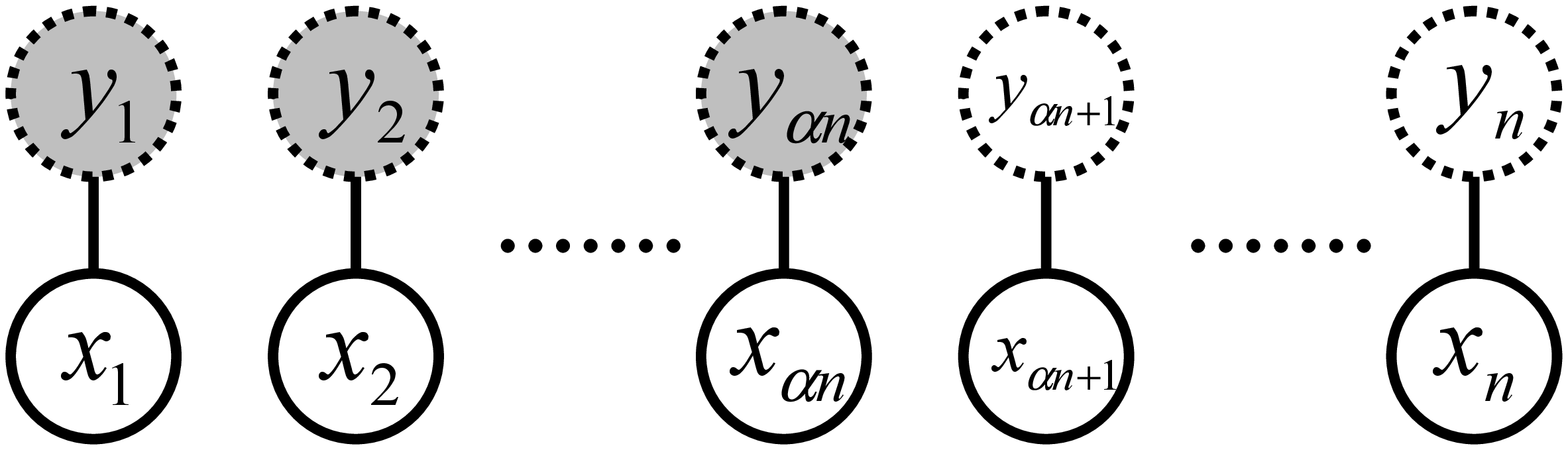}
\text{Type II'}
\label{fig:regfig5}
\end{minipage}
&
\begin{minipage}{0.5\hsize}
\centering
\includegraphics[angle=-90,width=\columnwidth]{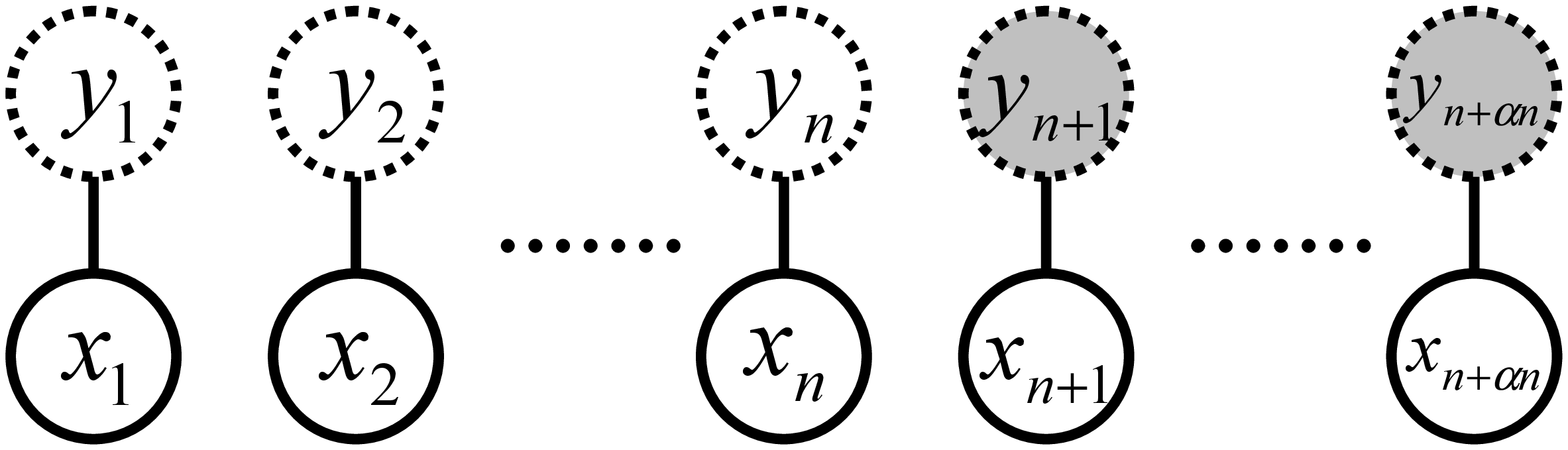}
\text{Type III'}
\label{fig:regfig6}
\end{minipage}
\end{tabular}
\caption{(Left) Partial marginal estimation for $y_1,\dots,y_{\alpha n}$.
(Right) Estimation for future data $y_{n+1},\dots,y_{n+\alpha n}$.}
\label{fig:vTypes}
\end{figure}
Consider the variants of Types II and III depicted in Figure \ref{fig:vTypes}.
Assume that $0<\alpha\le 1$ is a constant rational number
and that $n$ gets large enough to satisfy that $\alpha n$ is an integer.
The left panel shows the partial marginal estimation referred to as Type II'.
We will consider the joint probability of $y_1,\dots, y_{\alpha n}$, 
where the remaining variables $y_{\alpha n+1},\dots,y_n$ have been marginalized out.
Type II' is equivalent to Type I when $\alpha=1$.
Note that the order in which the target nodes are determined
does not change the average accuracy for i.i.d. data.
The right panel indicates the estimations for future data $y_{n+1},\dots,y_{n+\alpha n}$.
We refer to it as Type III' and construct the joint probability on these variables.
In the variant types, the targets are changed from a single node to $\alpha n$ nodes,
which enables us to analyze the asymptotic behavior.

We will use the following notation:
\begin{align*}
X_1 =& \{x_1,\dots,x_{\alpha n}\},\\
Y_1 =& \{y_1,\dots,y_{\alpha n}\}
\end{align*}
for Type II' and
\begin{align*}
X_2 =& \{x_{n+1},\dots,x_{n+\alpha n}\},\\
Y_2 =& \{y_{n+1},\dots,y_{n+\alpha n}\}
\end{align*}
for Type III'.
The Bayes estimations are given by
\begin{align*}
p(Y_1|X^n) =& \frac{\int \prod_{j=1}^{\alpha n} p(x_j,y_j|w)\prod_{i=\alpha n+1}^n p(x_i|w)\varphi(w;\eta)dw}
{\int \prod_{i=1}^n p(x_i|w)\varphi(w;\eta)dw},\\
p(Y_2|X_2,X^n) =& \int \prod_{i=n+1}^{n+\alpha n}\frac{p(x_i,y_i|w)}{p(x_i|w)}p(w|X^n)dw
\end{align*}
for Type II' and Type III', respectively.
The respective error functions are defined by
\begin{align*}
D_{Y_1|X^n}(n) =& \frac{1}{\alpha n}E_{X^n}\bigg[\sum_{Y_1}q(Y_1|X^n)\ln\frac{q(Y_1|X^n)}{p(Y_1|X^n)}\bigg],\\
D_{Y_2|X_2}(n) =& \frac{1}{\alpha n}E_{X^n,X_2}\bigg[\sum_{Y_2}q(Y_2|X_2)\ln\frac{q(Y_2|X_2)}{p(Y_2|X_2,X^n)}\bigg].
\end{align*}
In ways similar to the proofs of Theorems \ref{th:asympMLerror} and \ref{th:asympBayeserror},
the asymptotic forms are derived as follows.
\begin{theorem}
\label{th:type2d}
In Type II', the error function has the following asymptotic form:
\begin{align*}
D_{Y_1|X^n}(n) =& \frac{1}{2\alpha n}\ln\det[K_{XY}(w^*)I_X(w^*)^{-1}] +o\bigg(\frac{1}{n}\bigg),
\end{align*}
where $K_{XY}(w)=\alpha I_{XY}(w)+(1-\alpha)I_X(w)$.
\end{theorem}
The proof is in the appendix.
\begin{theorem}
\label{th:type3d}
In Type III', the error function has the following asymptotic form:
\begin{align*}
D_{Y_2|X_2}(n) =& \frac{1}{2\alpha n}\ln\det[K_{XY}(w^*)I_X^{-1}(w^*)]+o\bigg(\frac{1}{n}\bigg).
\end{align*}
\end{theorem}
This proof is also in the appendix.
These theorems show that when Types II' and III' have the same $\alpha$, they asymptotically have the same accuracy.
This implies the asymptotic equivalency of Types II and III
by combining the results of the maximum likelihood method.

\begin{table}[t]
\centering
\caption{Coefficients of the dominant order $1/n$ in the error functions}
\begin{tabular}{|c|c|c|c|c|}
\hline
& Pred. & Type I & Type II' & Type III' \\
\hline
ML & $d/2$ & $\mathrm{Tr}[\{I_{XY}-I_X\}I_X^{-1}]/2$ & $\mathrm{Tr}[\{I_{XY}-I_X\}I_X^{-1}]/2$ & $\mathrm{Tr}[\{I_{XY}-I_X\}I_X^{-1}]/2$ \\
\hline
Bayes & $d/2$ & $\ln\det[I_{XY}I_X^{-1}]/2$ & $\ln\det[K_{XY}I_X^{-1}]/(2\alpha)$ & $\ln\det[K_{XY}I_X^{-1}]/(2\alpha)$ \\
\hline
\end{tabular}
\label{tab:errors_d}
\end{table}
Table \ref{tab:errors_d} summarizes the results.
Based on the definitions,
the results for the maximum likelihood method are also available for Types II' and III'.
Using the asymptotic forms,
we can compare the relation of the magnitudes for the maximum likelihood method.
\begin{corollary}
\label{cor:comp_Bayes2ML}
Assume that $I_{XY}(w)\neq I_X(w)$.
For $0< \alpha \le 1$, there exists a positive constant $c_1$ such that
\begin{align*}
\mathrm{Tr}[\{I_{XY}(w)-I_X(w)\}I_X^{-1}(w)] - \frac{1}{\alpha}\ln \det [K_{XY}(w)I_X^{-1}(w)]
\ge \frac{c_1}{n} + o\bigg(\frac{1}{n}\bigg).
\end{align*}
\end{corollary}
The proof is in the appendix.
We immediately obtain the following relation,
which shows the advantage of the Bayes estimation in the asymptotic case:
\begin{align*}
D_{Y_1|X^n}^{\text{Bayes}}(n) <& D_{Y_1|X^n}^{\text{ML}}(n)\\
D_{Y_2|X_2}^{\text{Bayes}}(n) <& D_{Y_2|X_2}^{\text{ML}}(n)
\end{align*}
for respective $\alpha$'s.

By comparing the errors of Types I and II' in the Bayes method,
we can obtain the effect of supplementary observable data.
Let us consider the Type-II' case in which the estimation target is $Y_1$
and the training data is only $X_1$.
This corresponds to the estimation in Type I with $\alpha n$ training data, which we emphasize by calling it Type I'.
The difference between Type I' and Type II' is the addition of supplementary data $X^n\setminus X_1$.
\begin{corollary}
\label{cor:supplement}
Assume that the minimum eigenvalue of $I_{XY}(w^*)I_X^{-1}(w^*)$ is not less than one,
i.e., $\lambda_d \ge 1$.
The error difference is asymptotically described as
\begin{align*}
D(\alpha n)-D_{Y_1|X^n}(n) =& \frac{1}{2\alpha n}\ln \det [I_{XY}(w^*)K_{XY}^{-1}(w^*)] +o\bigg(\frac{1}{n}\bigg)\\
\ge& \frac{c_2}{n} + o\bigg(\frac{1}{n}\bigg),
\end{align*}
where $c_2$ is a positive constant.
This shows that Type II' has a smaller error than Type I' in the asymptotic situation;
the supplementary data make the estimation more accurate.
\end{corollary}
The proof is in the appendix.
\subsection{Comparison between the Two Methods}
Corollaries \ref{cor:comp} and \ref{cor:comp_Bayes2ML} show 
that the Bayes method is more accurate than the maximum likelihood method for Types I, II', and III'.
There have been many data-based comparisons of the predicting performances of these two methods
(e.g., \cite{Akaike80, Mackay1992, Draper2005}).
We will now discuss the computational costs of the two methods for the estimation of latent variables.  
We note there will be a trade-off between cost and accuracy.

We will assume that the estimated distribution is to be calculated for a practical purpose.
For example, the value of $p(Y^n|X^n)$ in Type I is used for sampling label assignments
and for searching for the optimal assignment $\arg\max_{Y^n} p(Y^n|X^n)$.
The maximum likelihood method requires the determination of $\hat{w}_X$ for all Types I, II, and III.
The computation is not expensive once $\hat{w}_X$ is successfully found, 
but the global maximum point of the likelihood function is not easily obtained.
The EM algorithm is commonly used for searching for the maximum likelihood estimator in models with latent variables, 
but it is often trapped in one of the local maxima.
The results of the steepest descent method also depend on the initial point and the step size of the iteration.

The Bayes method is generally expensive.
In the estimated distribution $p(Y^n|X^n)$ of Type I,
the numerator $Z(X^n,Y^n)$ contains integrals that depend on $Y^n$.
Sampling $y_i$ in Type II requires the same computation as for Type I:
we can obtain $y_i$ by ignoring the other elements $Y^n\setminus y_i$,
which realizes the marginalization $\sum_{Y^n\setminus y_i}p(Y^n|X^n)$.
A conjugate prior allows us to have a tractable form of $Z(X^n,Y^n)$ \cite{Dawid1993,Heckerman1999},
which reduces the computational cost.
In Type III, Eq.\ref{eq:BayesTypeIII} shows that there is no direct sampling method for $y$.
In this case, expensive sampling from the posterior $p(w|X^n)$ is necessary.

The VB method is an approximation that allows the direct computation of $P(Y^n|X^n)$ and $p(w|X^n)$,
which have tractable forms and reduced computational costs.
However, the assumption that $P(Y^n|X^n)$ and $p(w|X^n)$ are independent does not hold in many cases.
We conjecture that the $P(Y^n|X^n)$ of the VB method
will be less accurate than that of the original Bayes method.
\section{Conclusions}
\label{sec:conc}
In the present paper we formalized the estimation from the observable data of the distribution 
of the latent variables,
and we measured its accuracy by using the Kullback-Leibler divergence.
We succeeded in deriving the asymptotic error functions
for both the maximum likelihood and the Bayes methods.
These results allow us to mathematically compare the estimation methods:
we determined that the Bayes method is more accurate than
the maximum likelihood method in most cases,
while their prediction accuracies are equivalent.
The generalization error has been approximated from the given observable data,
such as by using the cross-validation and bootstrap methods, 
but there is no approximation technique for the error of the estimation of the latent variables,
because the latent data can not be obtained.
Therefore, these asymptotic forms are thus far the only way we have to estimate
their accuracy.

\section*{Acknowledgement}
This research was partially supported by the Kayamori Foundation of Informational Science Advancement
and KAKENHI 23500172.
%
%
%
%

\appendix
\section*{Appendix}
In this section, we prove the theorems and the corollaries.
\subsection*{Proof of Theorem \ref{th:asympMLerror}}
%
First, let us define another Fisher information matrix:
\begin{align*}
\{ I_{Y|X}(w) \}_{ij} =& E\biggl[ \frac{\partial \ln p(y|x,w)}{\partial w_i}
\frac{\partial \ln p(y|x,w)}{\partial w_j}\biggr].
\end{align*}
Based on $p(y|x,w)=p(x,y|w)/p(x|w)$,
\begin{align*}
I_{Y|X}(w) =& I_{XY}(w)+I_X(w)-J_{XY}(w)-J_{XY}^\top(w),
\end{align*}
where
\begin{align*}
\{ J_{XY}(w) \}_{ij} =& E\biggl[ \frac{\partial \ln p(x,y|w)}{\partial w_i}
\frac{\partial \ln p(x|w)}{\partial w_j}\biggr].
\end{align*}
According to the definition, we obtain
\begin{align*}
\{J_{XY}(w) \}_{ij} =& E\bigg[ \frac{1}{p(x,y|w)}\frac{\partial p(x,y|w)}{\partial w_i}
\frac{\partial \ln p(x|w)}{\partial w_j}\bigg]\\
=&\int \sum_y \frac{\partial p(x,y|w)}{\partial w_i}
\frac{\partial \ln p(x|w)}{\partial w_j}dx \\
=&\int \frac{\partial p(x|w)}{\partial w_i}
\frac{\partial \ln p(x|w)}{\partial w_j}dx \\
=&\int \frac{\partial \ln p(x|w)}{\partial w_i}
\frac{\partial \ln p(x|w)}{\partial w_j} p(x|w)dx =\{ I_X(w) \}_{ij}.
\end{align*}
Thus, it holds that
\begin{align}
I_{Y|X}(w) =& I_{XY}(w) - I_X(w).\label{eq:equivIs}
\end{align}

Next, let us divide the error function into three parts:
\begin{align}
D(n) =& D_1(n) - D_2(n) - D_3(n),\label{eq:D123}\\
D_1(n) =& \frac{1}{n}E_{X^nY^n}\big[\ln q(X^n,Y^n)\big],\nonumber\\
D_2(n) =& \frac{1}{n}E_{X^nY^n}\big[\ln p(X^n,Y^n|\hat{w}_X)\big],\nonumber\\
D_3(n) =& \frac{1}{n}E_{X^n}\biggl[ \ln\frac{q(X^n)}{p(X^n|\hat{w}_X)}\biggr],\nonumber
\end{align}
where the expectation is
\begin{align*}
E_{X^nY^n}[f(X^n,Y^n)]=\int \sum_{Y^n} f(X^n,Y^n)q(X^n,Y^n)dX^n.
\end{align*}
Because $D_3(n)$ is the training error on $p(x|\hat{w}_X)$,
the asymptotic form is known \cite{Akaike}:
\begin{align*}
D_3(n) =& -\frac{d}{2n} + o\bigg(\frac{1}{n}\bigg).
\end{align*}
Let another estimator be defined by
\begin{align*}
\hat{w}_{XY} =& \arg\max L_{XY}(w).
\end{align*}
According to the Taylor expansion, $D_2(n)$ can be rewritten as
\begin{align}
D_2(n) =& \frac{1}{n}E_{X^nY^n}\bigg[ \sum_{i=1}^n \ln p(X_i,Y_i|\hat{w}_{XY})\bigg]\nonumber\\
&+\frac{1}{n}E_{X^nY^n}\bigg[ \delta w^\top\sum_{i=1}^n\frac{\partial\ln p(X_i,Y_i|\hat{w}_{XY})}{\partial w}\bigg]\nonumber\\
&+\frac{1}{2n}E_{X^nY^n}\bigg[ \delta w^\top\sum_{i=1}^n\frac{\partial^2\ln p(X_i,Y_i|\hat{w}_{XY})}{\partial w^2}\delta w 
+ R_1(\delta w)\bigg]\nonumber\\
=& \frac{1}{n}E_{X^nY^n}\bigg[ \sum_{i=1}^n \ln p(X_i,Y_i|\hat{w}_{XY})\bigg]\nonumber\\
&- \frac{1}{2}E_{X^nY^n}\big[ \delta w^\top I_{XY}(w^*) \delta w \big] +o\bigg(\frac{1}{n}\bigg),\nonumber
\end{align}
where $\delta w=\hat{w}_X-\hat{w}_{XY}$, and $R_1(\delta w)$ is the remainder term.
The matrix $\sum_{i=1}^n\frac{\partial^2\ln p(X_i,Y_i|\hat{w}_{XY})}{\partial w^2}$
was replaced with $I_{XY}(w^*)$ on the basis of the law of large numbers.
As for the first term of $D_2$,
\begin{align*}
&D_1(n)-\frac{1}{n}E_{X^nY^n}\bigg[ \sum_{i=1}^n \ln p(X_i,Y_i|\hat{w}_{XY})\bigg]\nonumber\\
&=-\frac{d}{2n} + o\bigg(\frac{1}{n}\bigg)
\end{align*}
because it is the training error on $p(x,y|\hat{w}_{XY})$.
The factor in the second term of $D_2$ can be rewritten as
\begin{align}
&E_{X^nY^n}\big[ \delta w^\top I_{XY}(w^*)\delta w \big] \nonumber\\
&= E_{X^nY^n}\big[ (\hat{w}_X-w^*)^\top I_{XY}(w^*)(\hat{w}_X-w^*) \big]\nonumber\\
&- E_{X^nY^n}\big[ (\hat{w}_{XY}-w^*)^\top I_{XY}(w^*)(\hat{w}_X-w^*) \big]\nonumber\\
&- E_{X^nY^n}\big[ (\hat{w}_X-w^*)^\top I_{XY}(w^*)(\hat{w}_{XY}-w^*) \big]\nonumber\\
&+ E_{X^nY^n}\big[ (\hat{w}_{XY}-w^*)^\top I_{XY}(w^*)(\hat{w}_{XY}-w^*) \big].\label{eq:est_cov}
\end{align}
Let us define an extended likelihood function,
\begin{align*}
L_2(w_{12}) =& \sum_{i=1}^n\ln p(X_i,Y_i|w_1) + \sum_{i=1}^n\ln p(X_i|w_2),
\end{align*}
where $w_{12}=(w_1^\top,w_2^\top)^\top$, $\hat{w}_{12}=(\hat{w}_{XY}^\top,\hat{w}_X^\top)^\top$,
and $w^{**}=(w^{*\top},w^{*\top})^\top$ are extended vectors.
According to the Taylor expansion,
\begin{align*}
\frac{\partial L_2(w_{12})}{\partial w_{12}} =& \bigg( \frac{\partial \sum \ln p(X_i,Y_i|w^*)}{\partial w_1}^\top,
\frac{\partial \sum \ln p(X_i|w^*)}{\partial w_2}^\top\bigg)^\top\nonumber\\
& -M \delta w_{12},\\
\delta w_{12} =& w_{12}-w^{**}\\
M =&
\begin{bmatrix}
-\frac{\partial^2 \sum \ln p(X_i,Y_i|w^*)}{\partial w_1^2} & 0 \\
0 & -\frac{\partial^2 \sum \ln p(X_i|w^*)}{\partial w_2^2}
\end{bmatrix}.
\end{align*}
According to $\frac{\partial L_2(\hat{w}_{12})}{\partial w_{12}}=0$,
$\delta\hat{w}_{12}=\hat{w}_{12}-w^{**}$ can be written as
\begin{align*}
\delta\hat{w}_{12} =& M^{-1}\bigg( \frac{\partial \sum \ln p(X_i,Y_i|w^*)}{\partial w_1}^\top,
\frac{\partial \sum \ln p(X_i|w^*)}{\partial w_2}^\top\bigg)^\top .
\end{align*}
Based on the central limit theorem,
$\delta\hat{w}_{12}$ is distributed from $\mathcal{N}(0,nM^{-1}\Sigma^{-1}M^{-1})$,
where
\begin{align*}
\Sigma^{-1} =&
\begin{bmatrix}
I_{XY}(w^*) & J_{XY}(w^*)\\
J_{XY}^\top(w^*) & I_X(w^*)
\end{bmatrix}.
\end{align*}
The covariance $nM^{-1}\Sigma^{-1}M^{-1}$ of $\delta \hat{w}_{12}$
directly shows the covariance of the estimators $\hat{w}_X$ and $\hat{w}_{XY}$
in Eq.\ref{eq:est_cov}.
Thus it holds that
\begin{align*}
&E_{X^nY^n}\big[ \delta w^\top I_{XY}(w^*)\delta w \big] \nonumber\\
&= \frac{1}{n}\mathrm{Tr}\bigg[ I_{XY}(w^*)I_X^{-1}(w^*)\bigg]
-\frac{1}{n}\mathrm{Tr}\bigg[ J_{XY}(w^*)I_X^{-1}(w^*)\bigg]\nonumber\\
&-\frac{1}{n}\mathrm{Tr}\bigg[ J_{XY}^\top(w^*)I_X^{-1}(w^*)\bigg]
+\frac{1}{n}\mathrm{Tr}\bigg[ I_X(w^*)I_X^{-1}(w^*)\bigg] + o\bigg(\frac{1}{n}\bigg).
\end{align*}
Considering the relation Eq.\ref{eq:D123},
we obtain that
\begin{align*}
D(n) =& \frac{1}{2n}\mathrm{Tr}[I_{Y|X}(w^*)I_X^{-1}(w^*)] +o\bigg(\frac{1}{n}\bigg).
\end{align*}
Based on Eq.\ref{eq:equivIs}, the theorem is proved.
{\bf (End of Proof)}
\subsection*{Proof of Theorem \ref{th:asympBayeserror}}
%
Let us define the following entropy functions:
\begin{align*}
S_{XY} = -\sum_{y=1}^{K^*}\int q(x,y)\ln q(x,y) dx,\\
S_{X} =-\int q(x)\ln q(x) dx.
\end{align*}
According to the definition,
the error function Eq.\ref{eq:deferror} with the Bayes estimation can be rewritten as
\begin{align*}
D(n) =& \frac{1}{n}\bigg\{F_{XY}(n) - F_X(n)\bigg\},
\end{align*}
where
\begin{align*}
F_{XY}(n) =& -nS_{XY} -E_{X^nY^n}\bigg[ \ln Z(X^n,Y^n)\bigg],\\
F_X(n) =& -nS_X -E_{X^n}\bigg[ \ln Z(X^n) \bigg].
\end{align*}
Based on the Taylor expansion at $w=\hat{w}_X$,
\begin{align*}
F_X(n) =& -nS_X -E_{X^n}\bigg[ \ln\int\exp\bigg\{\ln p(X^n|\hat{w}_X)\nonumber\\
&+\frac{1}{2}(w-\hat{w}_X)^\top \frac{\partial^2 \ln p(X^n|\hat{w}_X)}{\partial w^2} (w-\hat{w}_X)+r_1(w)\bigg\}\varphi(w;\eta)dw\bigg]\\
=& -nS_X -E_{X^n}[\ln p(X^n|\hat{w}_X] -E_{X^n}\bigg[ \ln\int e^{r_1(w)}\varphi(w;\eta)\mathcal{N}(\hat{w}_X,\Sigma_1/n)dw\bigg],
\end{align*}
where $r_1(w)$ is the remainder term and 
\begin{align*}
\Sigma_1^{-1} =& - \frac{1}{n}\frac{\partial^2 \ln p(X^n|\hat{w}_X)}{\partial w^2},
\end{align*}
which converges to $I_X(w^*)$ based on the law of large numbers.
Again, applying the expansion at $w=w^*$ to $e^{r_1(w)}\varphi(w;\eta)$, we obtain
\begin{align*}
F_X(n) =& E_{X^n}\bigg[\ln\frac{q(X^n)}{p(X^n|\hat{w}_X)}\bigg]-\ln \sqrt{2\pi}^d \sqrt{\det\{nI_X(w^*)\}^{-1}}\nonumber\\
&-E_{X^n}\bigg[\ln\int\bigg\{e^{r_1(w^*)}\varphi(w^*:\eta)\nonumber\\
&+(w-w^*)^\top\frac{\partial e^{r_1(w^*)}\varphi(w^*;\eta)}{\partial w}+r_2(w)\bigg\}
\mathcal{N}\big(\hat{w}_X,\big\{nI_X(w^*)\big\}^{-1}\big)dw\bigg] + o(1),
\end{align*}
where $r_2(w)$ is the remainder term.
The first term is the training error on $p(x|\hat{w}_X)$.
According to \cite{Akaike}, it holds that
\begin{align*}
E_{X^n}\bigg[\ln\frac{q(X^n)}{p(X^n|\hat{w}_X)}\bigg]=&-\frac{d}{2} +o(1).
\end{align*}
Then, we obtain
\begin{align*}
F_X(n) =& \frac{d}{2}\ln\frac{n}{2\pi e}+\ln\frac{\sqrt{\det I_X(w^*)}}{\varphi(w^*;\eta)}+o(1),
\end{align*}
which is consistent with the result of \cite{Clarke90}.
By replacing $X^n$ with $(X^n,Y^n)$,
\begin{align*}
F_{XY}(n) =& \frac{d}{2}\ln\frac{n}{2\pi e}+\ln\frac{\sqrt{\det I_{XY}(w^*)}}{\varphi(w^*;\eta)}+o(1) .
\end{align*}
Therefore,
\begin{align*}
D(n) =& \frac{1}{2n}\bigg\{ \ln\det I_{XY}(w^*)-\ln\det I_X(w^*)\bigg\} +o\bigg(\frac{1}{n}\bigg),
\end{align*}
which proves the theorem.
{\bf (End of Proof)}
\subsection*{Proof of Corollary \ref{cor:comp}}
%
Because $I_{XY}(w)$ is symmetric positive definite, 
we have a decomposition $I_{XY}(w)=LL^\top$, where $L$ is a lower triangular matrix.
The other Fisher information matrix $I_X(w)$ is also symmetric positive definite.
Thus, $L^TI_X^{-1}(w)L$ is positive definite.
Let $\lambda_1\ge\lambda_2\ge\dots\ge\lambda_d>0$ be the eigenvalues of $L^\top I_X^{-1}(w)L$.
According to the assumption, at least one eigenvalue is different from the others.
Then, we obtain 
\begin{align*}
2n\{D^{\text{ML}}(n) - D^{\text{Bayes}}(n)\} =& \mathrm{Tr}[I_{XY}(w)I_X^{-1}(w)]-d -\ln\det[I_{XY}(w)I_X^{-1}(w)] +o(1)\\
=& \mathrm{Tr}[L^\top I_X^{-1}(w) L]-d-\ln\det[L^\top I_X^{-1}(w) L] +o(1)\\
=& \sum_{i=1}^d \{\lambda_i-1\} - \ln\prod_{i=1}^d \lambda_i+o(1)\\
=& \sum_{i=1}^d \{\lambda_i-1-\ln\lambda_i \} +o(1).
\end{align*}
The first term in the last expression is positive,
which proves the corollary.
{\bf (End of Proof)}
\subsection*{Proof of Theorem \ref{th:type2d}}
%
The error function is rewritten as
\begin{align*}
D_{Y_1|X^n}(n) =& \frac{1}{\alpha n}\bigg\{ F^{(1)}_{XY}(n) - F_X(n) \bigg\},\\
F^{(1)}_{XY}(n) =& -\alpha nS_{XY}-(1-\alpha)nS_X -E_{X^n,Y_1}\bigg[\ln \int L^{(1)}_{XY}(w)\varphi(w;\eta)dw\bigg],\\
L^{(1)}_{XY}(w) =& \prod_{j=1}^{\alpha n}p(x_j,y_j|w)\prod_{i=\alpha n+1}^n p(x_i|w).
\end{align*}
Based on the Taylor expansion at $w=\hat{w}^{(1)}$, where $\hat{w}^{(1)}=\arg\max L^{(1)}(w)$,
\begin{align*}
F^{(1)}_{XY}(n) =& E_{X^n,Y_1}\bigg[\sum_{j=1}^{\alpha n}\ln \frac{q(x_j,y_j)}{p(x_j,y_j|\hat{w}^{(1)})} 
+\sum_{i=\alpha n+1}^n \ln \frac{q(x_i)}{p(x_i|\hat{w}^{(1)})} \nonumber\\
&+ \ln\int\exp\bigg\{ -n(w-\hat{w}^{(1)})^\top G^{(1)}(X^n,Y_1)(w-\hat{w}^{(1)})
+r_3(w)\bigg\}\varphi(w;\eta)dw\bigg],
\end{align*}
where $r_3(w)$ is the remainder term and
\begin{align*}
G^{(1)}(X^n,Y_1) = -\frac{1}{n}\frac{\partial^2}{\partial w^2}
\bigg( \sum_{j=1}^{\alpha n} \ln p(x_j,y_j|\hat{w}^{(1)})+\sum_{i=\alpha n+1}^n \ln p(x_i|\hat{w}^{(1)})\bigg).
\end{align*}
The first and the second terms of $F^{(1)}_{XY}(n)$ correspond to the training error.
Following the same method as we used in the proof of Theorem \ref{th:asympBayeserror} and noting that
\begin{align*}
G^{(1)}(X^n,Y_1) \rightarrow K_{XY}(w^*),
\end{align*}
we obtain 
\begin{align*}
F^{(1)}_{XY}(n) =& \frac{d}{2}\ln\frac{n}{2\pi e}+\ln\frac{\sqrt{\det K_{XY}(w^*)}}{\varphi(w^*;\eta)} +o(1),
\end{align*}
which completes the proof.
{\bf (End of Proof)}
\subsection*{Proof of Theorem \ref{th:type3d}}
%
The error function is rewritten as
\begin{align*}
D_{Y_2|X_2}(n) =& \frac{1}{\alpha n}\bigg\{ F^{(2)}_{XY}(n) - F_X(n) \bigg\},\\
F^{(2)}_{XY}(n) =& -\alpha nS_{XY}-nS_X -E_{X^n,X_2,Y_2}\bigg[\ln \int L^{(2)}_{XY}(w)\varphi(w;\eta)dw\bigg],\\
L^{(2)}_{XY}(w) =& \prod_{j=n+1}^{n+\alpha n}p(y_j|x_j,w)\prod_{i=1}^n p(x_i|w).
\end{align*}
Based on the Taylor expansion at $w=\hat{w}^{(2)}$, where $\hat{w}^{(2)}=\arg\max L^{(2)}(w)$,
\begin{align*}
F^{(2)}_{XY}(n) =& E_{X^n,X_2,Y_2}\bigg[\sum_{j=n+1}^{\alpha n}\ln \frac{q(y_j|x_j)}{p(y_j|x_j,\hat{w}^{(2)})} 
+\sum_{i=1}^n \ln \frac{q(x_i)}{p(x_i|\hat{w}^{(2)})} \nonumber\\
&+ \ln\int\exp\bigg\{ -n(w-\hat{w}^{(2)})^\top G^{(2)}(X^n,X_2,Y_2)(w-\hat{w}^{(2)})
+r_4(w)\bigg\}\varphi(w;\eta)dw\bigg],
\end{align*}
where $r_4(w)$ is the remainder term and
\begin{align*}
G^{(2)}(X^n,X_2,Y_2) = -\frac{1}{n}\frac{\partial^2}{\partial w^2}
\bigg( \sum_{j=n+1}^{\alpha n} \ln p(y_j|x_j,\hat{w}^{(2)})+\sum_{i=1}^n \ln p(x_i|\hat{w}^{(2)})\bigg).
\end{align*}
The first and the second terms of $F^{(1)}_{XY}(n)$ correspond to the training error, which are stated as
\begin{align*}
E_{X^n,X_2,Y_2}\bigg[\sum_{j=n+1}^{\alpha n}\ln \frac{q(y_j|x_j)}{p(y_j|x_j,\hat{w}^{(2)})} 
+\sum_{i=1}^n \ln \frac{q(x_i)}{p(x_i|\hat{w}^{(2)})}\bigg] \nonumber\\
=-\mathrm{Tr}\bigg[\big\{\alpha I_{Y|X}(w^*)+I_X(w^*)\big\}K_{XY}(w^*)^{-1}\bigg] +o(1).
\end{align*}
Following the same method we used in the proof of Theorem \ref{th:asympBayeserror} and noting that
\begin{align*}
G^{(2)}(X^n,X_2,Y_2) \rightarrow K_{XY}(w^*),
\end{align*}
we obtain 
\begin{align*}
F^{(1)}_{XY}(n) =& -\mathrm{Tr}\bigg[\big\{\alpha I_{Y|X}(w^*)+I_X(w^*)\big\}K_{XY}(w^*)^{-1}\bigg]\nonumber\\
&+\frac{d}{2}\ln\frac{n}{2\pi}+\ln\frac{\sqrt{\det K_{XY}(w^*)}}{\varphi(w^*;\eta)} +o(1)\\
=& \frac{d}{2}\ln\frac{n}{2\pi e}+\ln\frac{\sqrt{\det K_{XY}(w^*)}}{\varphi(w^*;\eta)} +o(1),
\end{align*}
which completes the proof.
{\bf (End of Proof)}
\subsection*{Proof of Corollary \ref{cor:comp_Bayes2ML}}
%
It holds that
\begin{align*}
\frac{1}{\alpha}\ln \det[K_{XY}(w)I_X^{-1}(w)] =& \frac{1}{\alpha}\ln \det[\alpha\{I_{XY}(w)-I_X(w)\}I_X^{-1}(w)+E_d],
\end{align*}
where $E_d$ is the $d\times d$ unit matrix.
On the other hand,
\begin{align*}
\mathrm{Tr}[\{I_{XY}(w)-I_X(w)\}I_X^{-1}(w)] =& \frac{1}{\alpha}\bigg\{ \mathrm{Tr}[\alpha\{I_{XY}(w)-I_X(w)\}I_X^{-1}(w)+E_d]-d \bigg\}.
\end{align*}
It is easy to confirm that $\alpha L_1^\top I_X^{-1}(w)L_1+E_d$ is positive definite,
where $L_1^\top L_1=I_{XY}(w)-I_X(w)$.
Considering the eigenvalues $\mu_1 \ge \mu_2 \ge \dots \ge \mu_d>0$, we can obtain the following relation
in the same way as we did in the proof of Corollary \ref{cor:comp}:
\begin{align*}
\mathrm{Tr}[\{I_{XY}(w)-I_X(w)\}I_X^{-1}(w)] - \frac{1}{\alpha}\ln \det[K_{XY}(w)I_X^{-1}(w)]
=& \frac{1}{\alpha}\sum_{i=1}^d \bigg\{\mu_i -1 - \ln \mu_i\bigg\}.
\end{align*}
It is easy to confirm that the right-hand side is positive,
which completes the proof.
{\bf (End of Proof)}
\subsection*{Proof of Corollary \ref{cor:supplement}}
%
Based on the eigenvalues of $I_{XY}(w^*)I_X^{-1}(w^*)$, it holds that
\begin{align*}
\ln \det [I_{XY}(w^*)K_{XY}^{-1}(w^*)] =& \ln \det [I_{XY}(w^*)I_X^{-1}(w^*)]
-\ln \det[\alpha I_{XY}(w^*)I_X^{-1}(w^*) +(1-\alpha)E_d]\\
=& \sum_{i=1}^d \ln \lambda_i - \sum_{i=1}^d\ln \{\alpha\lambda_i +(1-\alpha)\}\ge 0,
\end{align*}
which completes the proof.
{\bf (End of Proof)}
%


\vskip 0.2in
\bibliography{LearningTheory}
\bibliographystyle{mlapa}

\end{document}